\documentclass{article}

\usepackage{microtype}
\usepackage{graphicx}
\usepackage{subfigure}
\usepackage{caption}
\usepackage{amsmath}
\usepackage{amssymb}
\usepackage{booktabs} 
\usepackage{wrapfig}

\usepackage{hyperref}
\usepackage{algorithm2e}

\usepackage[accepted]{icml2019}

\newcommand{\minisection}[1]{\textbf{#1}\hspace{0.3em}}
\SetKwComment{Comment}{$\triangleright$\ }{}

\newcommand{\task}{\mathcal{T}}

\newcommand{\Policy}{\pi_{\theta}}
\newcommand{\Context}{C} 
\newcommand{\context}{\mathbf{c}} 
\newcommand{\Z}{Z}
\newcommand{\z}{\mathbf{z}}
\newcommand{\state}{\mathbf{s}}
\newcommand{\act}{\mathbf{a}}
\newcommand{\E}{\mathbb{E}}

\icmltitlerunning{Efficient Off-Policy Meta-RL}

\begin{document}

\twocolumn[
\icmltitle{Efficient Off-Policy Meta-Reinforcement Learning via \\ Probabilistic Context Variables}

\icmlsetsymbol{equal}{*}

\begin{icmlauthorlist}
\icmlauthor{Kate Rakelly}{be,equal}
\icmlauthor{Aurick Zhou}{be,equal}
\icmlauthor{Deirdre Quillen}{be}
\icmlauthor{Chelsea Finn}{be}
\icmlauthor{Sergey Levine}{be}

\end{icmlauthorlist}

\icmlaffiliation{be}{EECS Department, UC Berkeley, Berkeley, CA, USA}

\icmlcorrespondingauthor{Kate Rakelly}{rakelly@eecs.berkeley.edu}

\icmlkeywords{meta-learning, reinforcement learning}

\vskip 0.3in
]



\printAffiliationsAndNotice{\icmlEqualContribution} 

\begin{abstract}
Deep reinforcement learning algorithms require large amounts of experience to learn an individual task. While in principle meta-reinforcement learning (meta-RL) algorithms enable agents to learn new skills from small amounts of experience, several major challenges preclude their practicality. Current methods rely heavily on on-policy experience, limiting their sample efficiency. The also lack mechanisms to reason about task uncertainty when adapting to new tasks, limiting their effectiveness in sparse reward problems. In this paper, we address these challenges by developing an off-policy meta-RL algorithm that disentangles task inference and control. In our approach, we perform online probabilistic filtering of latent task variables to infer how to solve a new task from small amounts of experience. This probabilistic interpretation enables posterior sampling for structured and efficient exploration. 
We demonstrate how to integrate these task variables with off-policy RL algorithms to achieve both meta-training and adaptation efficiency. Our method outperforms prior algorithms in sample efficiency by 20-100X as well as in asymptotic performance on several meta-RL benchmarks.

\end{abstract}

\section{Introduction}
\label{sec:intro}
The combination of reinforcement learning (RL) with powerful non-linear function approximators has led to a wide range of advances in sequential decision making problems.
However, conventional RL methods learn a separate policy per task, each often requiring millions of interactions with the environment.
Learning large repertoires of behaviors with such methods quickly becomes prohibitive.
Fortunately, many of the problems we would like our autonomous agents to solve share common structure. 
For example screwing a cap on a bottle and turning a doorknob both involve grasping an object in the hand and rotating the wrist.
Exploiting this structure to learn new tasks more quickly remains an open and pressing topic.
Meta-learning methods learn this structure from experience by making use of large quantities of experience collected across a distribution of tasks.
Once learned, these methods can adapt quickly to new tasks given a small amount of experience.

While meta-learned policies adapt to new tasks with only a few trials, during training, they require massive amounts of data drawn from a large set of distinct tasks, exacerbating the problem of sample efficiency that plagues RL algorithms.
Most current meta-RL methods require on-policy data during both meta-training and adaptation~\cite{finn2017model, wang2016learning, duan2016rl, mishra2018simple, rothfuss2018promp, houthooft2018evolved}, which makes them exceedingly inefficient during meta-training.
However, making use of off-policy data for meta-RL poses new challenges. 
Meta-learning typically operates on the principle that meta-training time should match meta-test time - for example, an image classification meta-learner tested on classifying images from five examples should be meta-trained to take in sets of five examples and produce accurate predictions~\cite{vinyals2016matching}.
This makes it inherently difficult to meta-train a policy to adapt using off-policy data, which is systematically different from the data the policy would see when it explores (on-policy) in a new task at meta-test time.

In this paper, we tackle the problem of efficient off-policy meta-reinforcement learning. 
To achieve both meta-training efficiency and rapid adaptation, we propose an approach that integrates online inference of probabilistic context variables with existing off-policy RL algorithms.
Rapid adaptation requires reasoning about distributions: when exposed to a new task for the first time, the optimal meta-learned policy must carry out a stochastic exploration procedure to visit potentially rewarding states, as well as adapt to the task at hand~\cite{gupta2018meta}.
During meta-training, we learn a probabilistic encoder that accumulates the necessary statistics from past experience into the context variables that enable the policy to perform the task.
At meta-test time, when the agent is faced with an unseen task, the context variables can be sampled and held constant for the duration of an episode, enabling temporally-extended exploration.
The collected trajectories are used to update the posterior over the context variables, achieving rapid trajectory-level adaptation.
In effect, our method adapts by sampling ``task hypotheses,'' attempting those tasks, and then evaluating whether the hypotheses were correct or not.
Disentangling task inference from action makes our approach particularly amenable to off-policy meta-learning; the policy can be optimized with off-policy data while the probabilistic encoder is trained with on-policy data to minimize distribution mismatch between meta-train and meta-test.

The primary contribution of our work is an off-policy meta-RL algorithm called probabilistic embeddings for actor-critic RL (PEARL). Our method achieves excellent sample efficiency during meta-training, enables fast adaptation by accumulating experience online, and performs structured exploration by reasoning about uncertainty over tasks.
In our experimental evaluation, we demonstrate state-of-the-art results with 20-100X improvement in meta-training sample efficiency and substantial increases in asymptotic performance over prior state-of-the-art on six continuous control meta-learning environments.
We further examine how our model conducts structured exploration to adapt rapidly to new tasks in a 2-D navigation environment with sparse rewards.
Our open-source implementation of PEARL can be found at \url{https://github.com/katerakelly/oyster}. 
\section{Related Work}
\label{sec:rw}

\minisection{Meta-learning.}
Our work builds on the meta-learning framework \cite{schmidhuber1987evolutionary, bengio1990learning, thrun1998learning} in the context of reinforcement learning.
Recently, meta-RL methods have been developed for meta-learning dynamics models \cite{clavera2018learning, saemundsson2018meta} and policies \cite{finn2017model, duan2016rl, mishra2018simple} that can quickly adapt to new tasks.

Recurrent \cite{duan2016rl, wang2016learning} and recursive \cite{mishra2018simple} meta-RL methods adapt to new tasks by aggregating experience into a latent representation on which the policy is conditioned.
These approaches can be categorized into what we will call \emph{context-based} meta-RL methods, since a neural network is trained to take experience as input as a form of task-specific context.
Similarly, our approach can also be considered context-based; however, we represent task contexts with probabilistic latent variables, enabling reasoning over task uncertainty. 
Instead of using recurrence, we leverage the Markov property in our permutation-invariant encoder to aggregate experience, enabling fast optimization especially for long-horizon tasks while mitigating overfitting.
While prior work has studied methods that can train recurrent Q-functions with off-policy Q-learning methods, such methods have often been applied to much simpler tasks~\cite{heess2015memory}, and in discrete environments~\cite{hausknecht2015deep}. 
Indeed, our own experiments in Section~\ref{sec:results-ablate} demonstrate that straightforward incorporation of recurrent policies with off-policy learning is difficult. 
Contextual methods have also been applied to imitation learning by conditioning the policy on a learned embedding of a demonstration and optimizing with behavior cloning \cite{duan2017one, james2018task}.

In contrast to context-based methods, \emph{gradient-based} meta-RL methods learn from aggregated experience using policy gradients \cite{finn2017model,stadie2018some, rothfuss2018promp, xu2018learning}, meta-learned loss functions \cite{sung2017learning,houthooft2018evolved}, or hyperparameters \cite{xu2018meta}. 
These methods focus on on-policy meta-learning. 
We instead focus on meta-learning from off-policy data, which is non-trivial to do with methods based on policy gradients and evolutionary optimization algorithms.
Beyond substantial sample efficiency improvements, we also empirically find that our context-based method is able to reach higher asymptotic performance, in comparison to methods using policy gradients. 

Outside of RL, meta-learning methods for few-shot supervised learning problems have explored a wide variety of approaches and architectures \cite{santoro2016meta, vinyals2016matching, ravi2017optimization,oreshkin2018tadam}. 
Our permutation-invariant embedding function is inspired by the embedding function of prototypical networks
\cite{snell2017prototypical}.
While they use a distance metric in a learned, deterministic embedding space to classify new inputs, our embedding is probabilistic and is used to condition the behavior of an RL agent.
To our knowledge, no prior work has proposed this particular embedding function for meta-RL.

\minisection{Probabilistic meta-learning.}
Prior work has applied probabilistic models to meta-learning in both supervised and reinforcement learning domains.
Hierarchical Bayesian models have been used to model few-shot learning \cite{fe2003bayesian, tenenbaum1999bayesian}, including approaches that perform gradient-based adaptation \cite{grant2018recasting, yoon2018bayesian}.
For supervised learning, \citet{rusu2019meta, gordon2019meta, finn2018probabilistic} adapt model predictions using probabilistic latent task variables inferred via amortized approximate inference. 
We extend this idea to off-policy meta-RL. 
In the context of RL, \citet{hausman2018learning} also conditions the policy on inferred task variables, but the aim is to compose tasks via the embedding space, while we focus on rapid adaptation to new tasks.
While we infer task variables and explore via posterior sampling, MAESN \cite{gupta2018meta} adapts by optimizing the task variables with gradient descent and explores by sampling from the prior.

\minisection{Posterior sampling.}
In classical RL, posterior sampling \cite{strens2000bayesian,osband2013more} maintains a posterior over possible MDPs and enables temporally extended exploration by acting optimally according to a sampled MDP.
Our approach can be interpreted as a meta-learned variant of this method; probabilistic context captures the current uncertainty over the task, allowing the agent to explore in new tasks in a similarly structured manner.

\minisection{Partially observed MDPs.} 
Adaptation at test time in meta-RL can be viewed as a special case of RL in a POMDP \cite{kaelbling1998planning} by including the task as the unobserved part of the state. 
We use a variational approach related to \citet{igl2018deep}
to estimate belief over the task.
While they focus on solving general POMDPs, we leverage the additional structure imposed by the meta-learning problem to simplify inference, and use posterior sampling for exploration in new tasks. 
\section{Problem Statement}
\label{sec:problem}
Our approach is motivated by situations in which the agent can leverage varied experiences from previous tasks to adapt quickly to the new task at hand.
Sample efficiency is central to our problem statement, both in terms of the number of samples from previous experience (meta-training efficiency), and in the amount of experience required in the new task (adaptation efficiency).
To achieve meta-training efficiency, we leverage off-policy RL in our approach. 
Adaptation efficiency requires the agent to reason about its uncertainty over tasks, particularly in sparse reward settings.
To capture uncertainty in our belief over the task, we learn a probabilistic latent representation of prior experience.
We formalize the problem statement in this section, formulate our approach to adaptation as probabilistic inference in Section~\ref{sec:context}, and explain how our approach can be integrated with off-policy RL algorithms in Section~\ref{sec:off-policy}.

Similar to previous meta-RL formulations, we assume a distribution of tasks $p(\task)$, 
where each task is a Markov decision process (MDP), consisting of a set of states,
actions, a transition function, and a bounded reward function.
We assume that the transition and reward functions are unknown, but can be sampled by taking actions in the environment.
Formally, a task $\task = \{p(\state_0), p(\state_{t+1} | \state_{t}, \act_{t}), r(\state_{t}, \act_t)\}$ consists of an initial state distribution $p(\state_0)$, transition distribution $p(\state_{t+1} | \state_t, \act_t)$, and reward function $r(\state_t, \act_t)$.
Note that this problem definition encompasses task distributions with varying transition functions (e.g., robots with different dynamics) and varying reward functions (e.g., navigating to different locations).
Given a set of training tasks sampled from $p(\task)$, the meta-training process learns a policy that adapts to the task at hand by conditioning on the history of past transitions, which we refer to as \emph{context} $\context$.
Let $\context_n^\task = (\state_n, \act_n, r_n, \state_{n}')$ be one transition in the task $\task$ so that $\context_{1:N}^\task$ comprises the experience collected so far.
At test-time, the policy must adapt to a new task drawn from $p(\task)$.

\section{Probabilistic Latent Context}
\label{sec:context}
We capture knowledge about how the current task should be performed in a latent probabilistic context variable $\Z$, on which we condition the policy as $\Policy(\act | \state, \z)$ in order to adapt its behavior to the task.
Meta-training consists of leveraging data from a variety of training tasks to learn to infer the value of $\Z$ from a recent history of experience in the new task, as well as optimizing the policy to solve the task given samples from the posterior over $\Z$.
In this section we describe the structure of the meta-trained inference mechanism.
We address how meta-training can be performed with off-policy RL algorithms in Section~\ref{sec:off-policy}.

\subsection{Modeling and Learning Latent Contexts}
\label{sec:learning-context}
To enable adaptation, the latent context $\Z$ must encode salient information about the task.
Recall that $\context_{1:N}^\task$ comprises experience collected so far; throughout this section we will often write $c$ for simplicity.
We adopt an amortized variational inference approach \cite{kingma2013auto, rezende2014stochastic, alemi2016deep} to learn to infer $\Z$.
We train an \emph{inference network} $q_{\phi}(\z | \context)$, parameterized by $\phi$, that estimates the posterior $p(\z | \context)$.
In a generative approach, this can be achieved by optimizing $q_{\phi}(\z | \context)$ to reconstruct the MDP by learning a predictive models of reward and dynamics.
Alternatively, $q_{\phi}(\z | \context)$ can be optimized in a model-free manner to model the state-action value functions or to maximize returns through the policy over the distribution of tasks.
Assuming this objective to be a log-likelihood, the resulting variational lower bound is:
\begin{equation}
\E_{\task} [\E_{\z \sim q_\phi(\z | \context^\task)}[R(\task, \z)
+ \beta D_{\text{KL}}(q_\phi(\z | \context^\task) || p(\z))]]
\end{equation}
where $p(\z)$ is a unit Gaussian prior over $\Z$ and $R(\task, \z)$ could be a variety of objectives, as discussed above. 
The KL divergence term can also be interpreted as the result of a variational approximation to an information bottleneck \cite{alemi2016deep} that constrains the mutual information between $\Z$ and $\context$.
Intuitively, this bottleneck constrains $\z$ to contain only information from the context that is necessary to adapt to the task at hand, mitigating overfitting to training tasks.
While the parameters of $q_{\phi}$ are optimized during meta-training, at meta-test time the latent context for a new task is simply inferred from gathered experience.

\begin{figure}[t]
  \centering
    \includegraphics[width=1.0\linewidth]{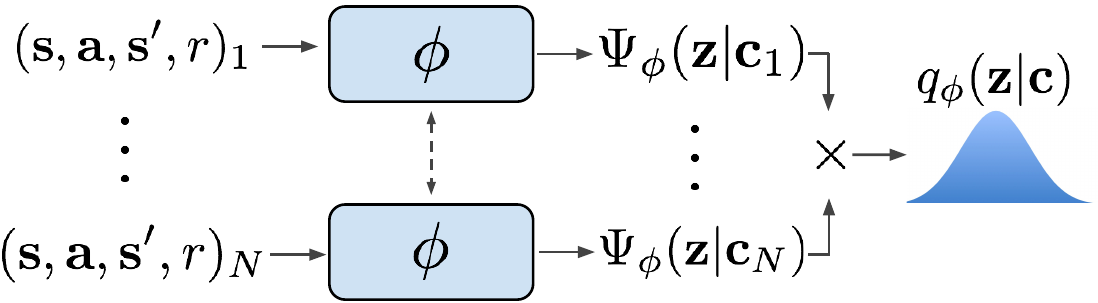}
  \caption{{\bf Inference network architecture}. The amortized inference network predicts the posterior over the latent context variables $q_{\phi}(\z | \context)$ as a permutation-invariant function of prior experience.
  \label{fig:encoder}
  }
\end{figure}

In designing the architecture of the inference network $q_{\phi}(\z | \context)$, we would like it to be expressive enough to capture minimal sufficient statistics of task-relevant information, without modeling irrelevant dependencies.
We note that an encoding of a fully observed MDP should be permutation invariant: if we would like to infer what the task is, identify the MDP model, or train a value function, it is enough to have access to a collection of transitions $\{\state_i, \act_i, \state'_i, r_i\}$, without regard for the order in which these transitions were observed.
With this observation in mind, we choose a permutation-invariant representation for $q_{\phi}(\z | \context_{1:N})$, modeling it as a product of independent factors 
\vspace{-1.0mm}
\begin{equation}
    q_\phi(\z | \context_{1:N}) \propto \Pi_{n=1}^N \Psi_\phi(\z | \context_n)
\end{equation}
\vspace{-0.5mm}
To keep the method tractable, we use Gaussian factors $\Psi_\phi(\z | \context_n) = \mathcal{N}(f_\phi^{\mu}(\context_n), f_\phi^{\sigma}(\context_n))$, which result in a Gaussian posterior.
The function $f_\phi$, represented as a neural network parameterized by $\phi$, predicts the mean $\mu$ as well as the variance $\sigma$ as a function of the  $\context_n$, is shown in Figure \ref{fig:encoder}.

\subsection{Posterior Sampling and Exploration via Latent Contexts}
\label{sec:explore}
Modeling the latent context as probabilistic allows us to make use of posterior sampling for efficient exploration at meta-test time.
In classical RL, posterior sampling \cite{strens2000bayesian, osband2013more} begins with a prior distribution over MDPs, computes a posterior distribution conditioned on the experience it has seen so far, and executes the optimal policy for a sampled MDP for the duration of an episode as an efficient method for exploration. 
In particular, acting optimally according to a random MDP allows for temporally extended (or deep) exploration, meaning that the agent can act to test hypotheses even when the results of actions are not immediately informative of the task.

In the single-task deep RL setting, posterior sampling and the benefits of deep exploration has been explored by \citet{osband2016bootstrapped},
which maintains an approximate posterior over value functions via bootstraps.
In contrast, our method PEARL directly infers a posterior over the latent context $\Z$, which may encode the MDP itself if optimized for reconstruction, optimal behaviors if optimized for the policy, or the value function if optimized for a critic.
Our meta-training procedure leverages training tasks to learn a prior over $\Z$ that captures the distribution over tasks and also learns to efficiently use experience to infer new tasks. 
At meta-test time, we initially sample $\z$'s from the prior and execute according to each $\z$ for an episode, thus exploring in a temporally extended and diverse manner.
We can then use the collected experience to update our posterior and continue exploring coherently in a manner that acts more and more optimally as our belief narrows, akin to posterior sampling.
\begin{figure}[t!]
  \centering
    \includegraphics[width=1.0\linewidth]{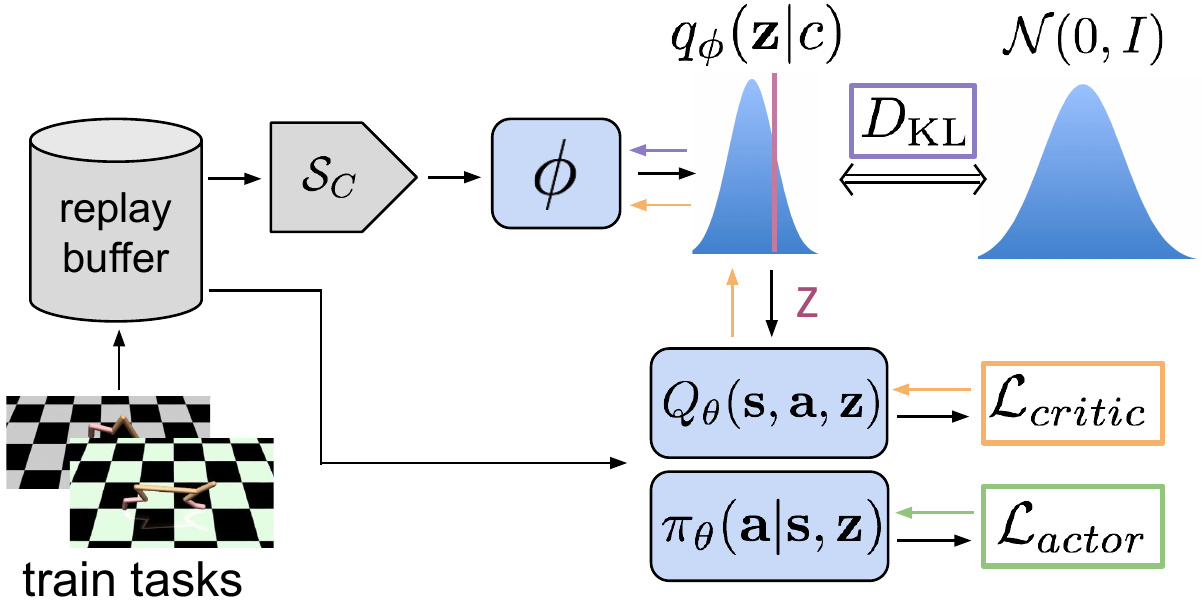}
  \caption{{\bf Meta-training procedure.} The inference network $q_{\phi}$ uses context data to infer the posterior over the latent context variable $\Z$, which conditions the actor and critic, and is optimized with gradients from the critic as well as from an information bottleneck on $\Z$. De-coupling the data sampling strategies for context ($\mathcal{S}_{\Context}$) and RL batches is important for off-policy learning.\vspace{-5mm}}
  \label{fig:train}
\end{figure}

\section{Off-Policy Meta-Reinforcement Learning}
\label{sec:off-policy}
While our probabilistic context model is straightforward to combine with on-policy policy gradient methods, a primary goal of our work is to enable efficient off-policy meta-reinforcement learning, where the number of samples for \emph{both} meta-training and fast adaptation is minimal. 
The efficiency of the meta-training process is largely disregarded in prior work, which make use of stable but relatively inefficient on-policy algorithms \cite{duan2016rl,finn2017model, gupta2018meta, mishra2018simple}.
However, designing off-policy meta-RL algorithms is non-trivial partly because modern meta-learning is predicated on the assumption that the distribution of data used for adaptation will match across meta-training and meta-test.
In RL, this implies that since at meta-test time on-policy data will be used to adapt, on-policy data should be used during meta-training as well. 
Furthermore, meta-RL requires the policy to reason about \emph{distributions}, so as to learn effective stochastic exploration strategies. 
This problem inherently cannot be solved by off-policy RL methods that minimize temporal-difference error, as they do not have the ability to directly optimize for distributions of states visited.
In contrast, policy gradient methods have direct control over the actions taken by the policy. 
Given these two challenges, a naive combination of meta-learning and value-based RL could be ineffective.
In practice, we were unable to optimize such a method.

\begin{algorithm}[t]
\begin{algorithmic}[1]
\REQUIRE Batch of training tasks $\{\task_i\}_{i=1 \ldots T}$ from $p(\task)$, learning rates $\alpha_1, \alpha_2, \alpha_3$
\STATE Initialize replay buffers $\mathcal{B}^i$ for each training task
\WHILE{not done}
\FOR{each $\task_i$} 
\STATE Initialize context $\context^{i} = \{\}$
\FOR{$k = 1, \ldots, K$}
\STATE Sample $\z \sim q_{\phi}(\z | \context^{i})$
\STATE Gather data from $\pi_{\theta}(\act | \state, \z)$ and add to $\mathcal{B}^{i}$
\STATE Update $\context^{i} = \{(\state_j, \act_j, \state'_j, r_j)\}_{j:1\ldots N} \sim \mathcal{B}^{i}$
\ENDFOR
\ENDFOR
\FOR{step in training steps}
\FOR{each $\task_i$}
\STATE Sample context $\context^i \sim \mathcal{S}_c(\mathcal{B}^i)$ and RL batch $b^i \sim \mathcal{B}^{i}$
\STATE Sample $\z \sim q_{\phi}(\z | \context^{i})$
\STATE $\mathcal{L}^i_{actor} = \mathcal{L}_{actor}(b^i, \z)$
\STATE $\mathcal{L}^i_{critic} = \mathcal{L}_{critic}(b^i, \z)$
\STATE $\mathcal{L}^i_{KL} = \beta D_{\text{KL}}(q(\z | \context^i) || r(\z))$
\ENDFOR
\STATE $\phi \gets \phi - \alpha_1 \nabla_\phi \sum_i \left(\mathcal{L}^i_{critic} + \mathcal{L}^i_{KL}\right)$
\STATE $\theta_{\pi} \gets \theta_{\pi} - \alpha_2  \nabla_\theta \sum_i \mathcal{L}^i_{actor}$
\STATE $\theta_{Q} \gets \theta_{Q} - \alpha_3  \nabla_\theta \sum_i \mathcal{L}^i_{critic}$
\ENDFOR
\ENDWHILE
\end{algorithmic}
\caption{{\bf PEARL Meta-training}}
\label{alg:meta-train}
\end{algorithm}

Our main insight in designing an off-policy meta-RL method with the probabilistic context in Section~\ref{sec:context} is that the data used to train the encoder need not be the same as the data used to train the policy.
The policy can treat the context $\z$ as part of the state in an off-policy RL loop, while the stochasticity of the exploration process is provided by the uncertainty in the encoder $q(\z|\context)$. 
The actor and critic are always trained with off-policy data sampled from the entire replay buffer $\mathcal{B}$.
We define a sampler $\mathcal{S}_{\context}$ to sample context batches for training the encoder.
Allowing $\mathcal{S}_{\context}$ to sample from the entire buffer presents too extreme of a distribution mismatch with on-policy test data.
However, the context does not need to be strictly on-policy; we find that an in-between strategy of sampling from a replay buffer of recently collected data retains on-policy performance with better efficiency.
We summarize our training procedure in Figure~\ref{fig:train} and Algorithm~\ref{alg:meta-train}.
Meta-testing is described in Algorithm~\ref{alg:meta-test}.

\subsection{Implementation}
We build our algorithm on top of the soft actor-critic algorithm (SAC) \cite{haarnoja2018soft}, an off-policy actor-critic method based on the maximum entropy RL objective which augments the traditional sum of discounted returns with the entropy of the policy.

SAC exhibits good sample efficiency and stability, and further has a probabilistic interpretation which integrates well with probabilistic latent contexts.
We optimize the parameters of the inference network $q(\z | \context)$ jointly with the parameters of the actor $\Policy(\act | \state, \z)$ and critic $Q_{\theta}(\state, \act, \z)$, using the reparameterization trick \cite{kingma2013auto} to compute gradients for parameters of $q_{\phi}(\z  | \context)$ through sampled $\z$'s.
We train the inference network using gradients from the Bellman update for the critic.
We found empirically that training the encoder to recover the state-action value function outperforms optimizing it to maximize actor returns, or reconstruct states and rewards.
The critic loss can then be written as,
\begin{equation}
\mathcal{L}_{critic} = \E_{\substack{(\state, \act, r, \state') \sim \mathcal{B} \\ \z \sim q_\phi(\z | \context)}} [Q_\theta(\state, \act, \z) - (r + \bar{V}(\state ', \bar{\z}))]^2 \\
\end{equation}
where $\bar{V}$ is a target network and $\bar{\z}$ indicates that gradients are not being computed through it. 
The actor loss is nearly identical to SAC, with the additional dependence on $\z$ as a policy input.
\begin{equation}
\mathcal{L}_{actor} \!=\! \E_{\substack{\state \sim \mathcal{B}, \act \sim \Policy \\ \z \sim q_\phi(\z | \context)}} \!\left[\! D_{\text{KL}}\!\left(\!\Policy(\act | \state, \bar{\z}) \! \left\| \frac{\text{exp}(Q_\theta(\state, \act, \bar{\z}))}{\mathcal{Z}_\theta(\state)} \! \right. \right) \! \right]
\end{equation}

Note that the context used to infer $q_\phi(\z | \context)$ is distinct from the data used to construct the critic loss.
As described in Section~\ref{sec:off-policy}, during meta-training we sample context batches separately from RL batches.
Concretely,the context data sampler $\mathcal{S}_{\context}$ samples uniformly from the most recently collected batch of data, recollected every $1000$ meta-training optimization steps.
The actor and critic are trained with batches of transitions drawn uniformly from the entire replay buffer.

\begin{algorithm}[t]
\begin{algorithmic}[1]
\REQUIRE test task $\task \sim p(\task)$
\STATE Initialize context $\context^{\task} = \{\}$
\FOR{$k=1, \ldots, K$}
\STATE Sample $z \sim q_{\phi}(\z | c^{\task})$
\STATE Roll out policy  $\pi_{\theta}(\act | \state, \z)$ to collect data $D^{\task}_{k} = \{(\state_j, \act_j, \state'_j, r_j)\}_{j:1\ldots N}$
\STATE Accumulate context $\context^{\task} = \context^{\task} \cup D^{\task}_{k}$
\ENDFOR
\end{algorithmic}
\caption{{\bf PEARL Meta-testing}}
\label{alg:meta-test}
\end{algorithm}
\begin{figure*}[t]
  \centering
    \includegraphics[width=1.04\textwidth]{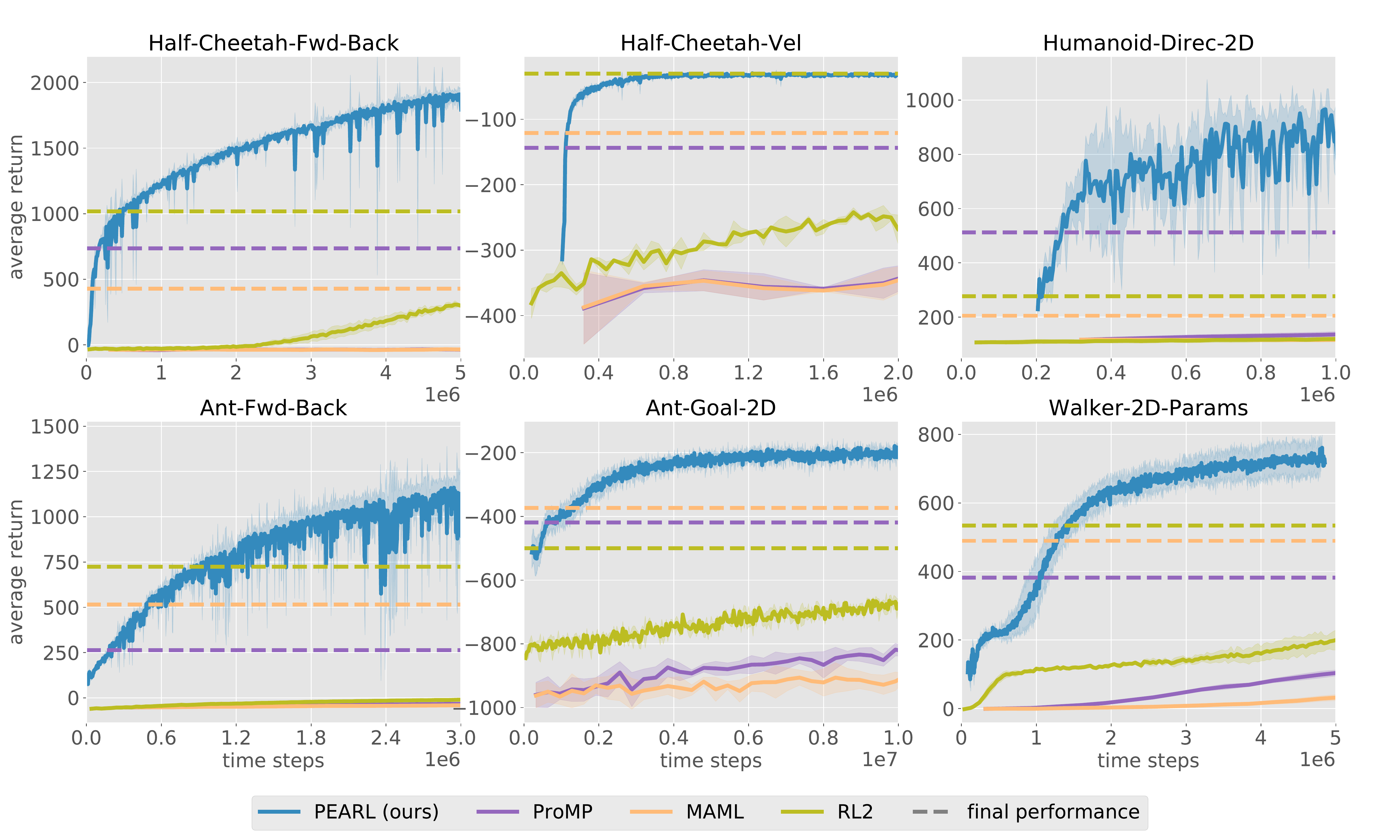}
  \caption{{\bf Meta-learning continuous control}. Test-task performance vs. samples collected during \emph{meta-training}. Our approach PEARL outperforms previous meta-RL methods both in terms of asymptotic performance and meta-training sample efficiency across six benchmark tasks. Dashed lines correspond to the maximum return achieved by each baseline after $1\text{e}8$ steps. By leveraging off-policy data during meta-training, PEARL is $20-100$x more sample efficient than the baselines, and achieves consistently better or equal final performance compared to the best performing prior method in each environment. See Appendix \ref{appendix:full-timescale} for the full timescale version of this plot.}
  \label{fig:core-results}
\end{figure*}

\section{Experiments}
\label{sec:experiments}
In our experiments, we assess the performance of our method and analyze its properties.
We first evaluate how our approach compares to prior meta-RL methods, especially in terms of sample efficiency, on several benchmark meta-RL problems in Section~\ref{sec:results-perf}.
We examine how probabilistic context and posterior sampling enable rapid adaptation via structured exploration strategies in sparse reward settings in Section~\ref{sec:results-explore}.
Finally, in Section~\ref{sec:results-ablate}, we evaluate the specific design choices in our algorithm through ablations.

\subsection{Sample Efficiency and Performance}
\label{sec:results-perf}
\minisection{Experimental setup.}
We evaluate PEARL on six continuous control environments focused around robotic locomotion, simulated via the MuJoCo simulator \cite{todorov2012mujoco}.
These locomotion task families require adaptation across reward functions (walking direction for Half-Cheetah-Fwd-Back, Ant-Fwd-Back, Humanoid-Direc-2D, target velocity for Half-Cheetah-Vel, and goal location for Ant-Goal-2D) or across dynamics (random system parameters for Walker-2D-Params).
These meta-RL benchmarks were previously introduced by \citet{finn2017model} and \citet{rothfuss2018promp}.
All tasks have horizon length $200$.
We compare to existing policy gradient meta-RL methods ProMP \cite{rothfuss2018promp} and MAML-TRPO \cite{finn2017model} using publicly available code.
We also re-implement the recurrence-based policy gradient RL$^2$ method \cite{duan2016rl} with PPO \cite{schulman2017proximal}.
The results of each algorithm are averaged across three random seeds.
We attempted to adapt recurrent DDPG \cite{heess2015memory} to our setting, but were unable to obtain reasonable results with this method.
We hypothesize that this is due to a combination of factors including the distribution mismatch in the adaptation data discussed in Section~\ref{sec:off-policy} and the difficulty of training with trajectories rather than decorrelated transitions.
This approach does not explicitly infer a belief over the task as we do, instead leaving the burden of both task inference and optimal behavior to the RNN.
In PEARL, decoupling task inference from the policy allows us the freedom to choose the encoder data and objective that work best with off-policy learning.
We experiment with recurrent architectures in the context of our own method in Section~\ref{sec:results-ablate}.

\minisection{Results.} 
To evaluate on the meta-testing tasks, we perform adaptation at the trajectory level, where the first trajectory is collected with context variable $\z$ sampled from the prior $r(\z)$.
Subsequent trajectories are collected with $\z \sim q(\z | \context)$ where the context is aggregated over all trajectories collected.
To compute final test-time performance, we report the average returns of trajectories collected after two trajectories have been aggregated into the context.
Notably, we find RL$^2$ to perform much better on these benchmarks than previously reported, possibly due to using PPO for optimization and selecting better hyper-parameters. We observe that
PEARL significantly outperforms prior meta-RL methods across all domains in terms of both asymptotic performance and sample efficiency, as shown in Figure~\ref{fig:core-results}. 
\begin{figure}[t]
  \centering
    \includegraphics[width=1.0\linewidth]{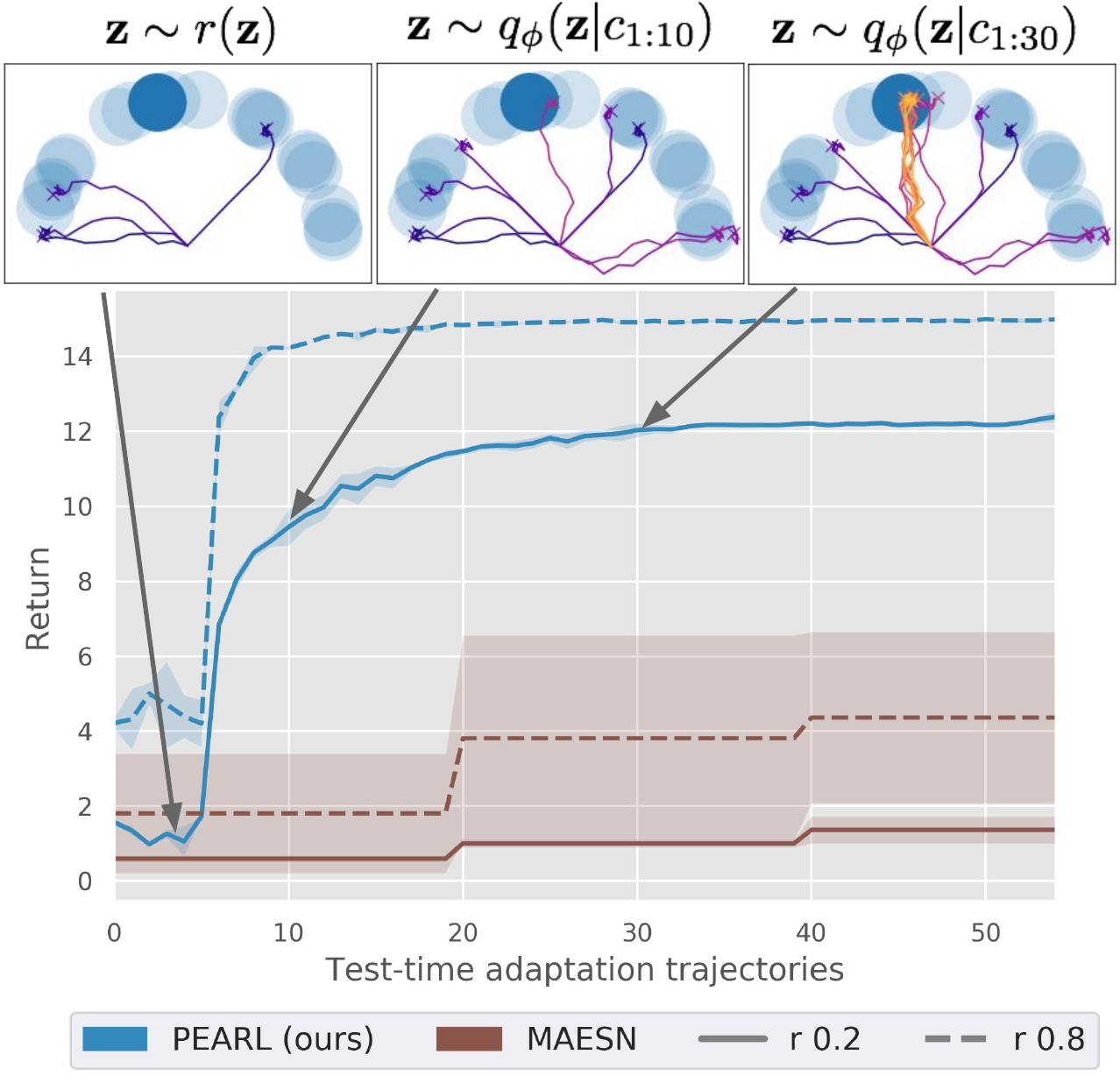}
    \caption{{\bf Sparse 2D navigation}. The agent must navigate to a previously unseen goal (dark blue, other test goals in light blue) with reward given only when inside the goal radius -- radius of 0.2 (illustrated) and 0.8 are tested here. The agent is trained to navigate to a training set of goals, then tested on a distinct set of unseen test goals. By using posterior sampling to explore efficiently, PEARL is able to start adapting to the task after collecting on average only 5 trajectories, outperforming MAESN~\cite{gupta2018meta}.\vspace{-5mm}}
  \label{fig:maesn-comparison}
\end{figure}
Here we truncate the $x$-axis at the number of timesteps required for PEARL to converge; see Appendix \ref{appendix:full-timescale} for the full timescale version of this plot.
We find that PEARL uses $20$-$100$x fewer samples during meta-training than previous meta-RL approaches while improving final asymptotic performance by 50-100\% in five of the six domains.

\subsection{Posterior Sampling For Exploration}
\label{sec:results-explore}
In this section we evaluate whether posterior sampling in our model enables effective exploration strategies in sparse reward MDPs.
Intuitively, by sampling from the prior context distribution $r(\z)$, the agent samples a hypothesis according to the distribution of training tasks it has seen before.
As the agent acts in the environment, the context posterior $p(\z | \context)$ is updated, allowing it to reason over multiple hypotheses to determine the task.
We demonstrate this behavior with a 2-D navigation task in which a point robot must navigate to different goal locations on edge of a semi-circle.
We sample training and testing sets of tasks, each consisting of $100$ randomly sampled goals. 
A reward is given only when the agent is within a certain radius of the goal.
We experiment with radius 0.2 and 0.8.
While our aim is to adapt to new tasks with sparse rewards, meta-training with sparse rewards is extremely difficult as it amounts to solving many sparse reward tasks from scratch.
For simplicity we therefore assume access to the dense reward during meta-training, as done by~\citet{gupta2018meta}, but this burden could also be mitigated with task-agnostic exploration strategies.

In this setting, we compare to MAESN \cite{gupta2018meta}, a prior method that also models probabilistic task variables and performs on-policy gradient-based meta-learning.
We demonstrate we are able to adapt to the new sparse goal in fewer trajectories.
Even with fewer samples, PEARL also outperforms MAESN in terms of final performance.
In Figure~\ref{fig:maesn-comparison} we compare adaptation performance on test tasks.
In addition to achieving higher returns and adapting faster, PEARL is also more efficient during meta-training.
Our results were achieved with $\sim 1e6$ timesteps while MAESN uses $\sim 1e8$ timesteps.

\subsection{Ablations}
\label{sec:results-ablate}
In this section we ablate the features of our approach to better understand the salient features of our method.
\begin{figure}
  \centering
    \includegraphics[width=0.5\textwidth]{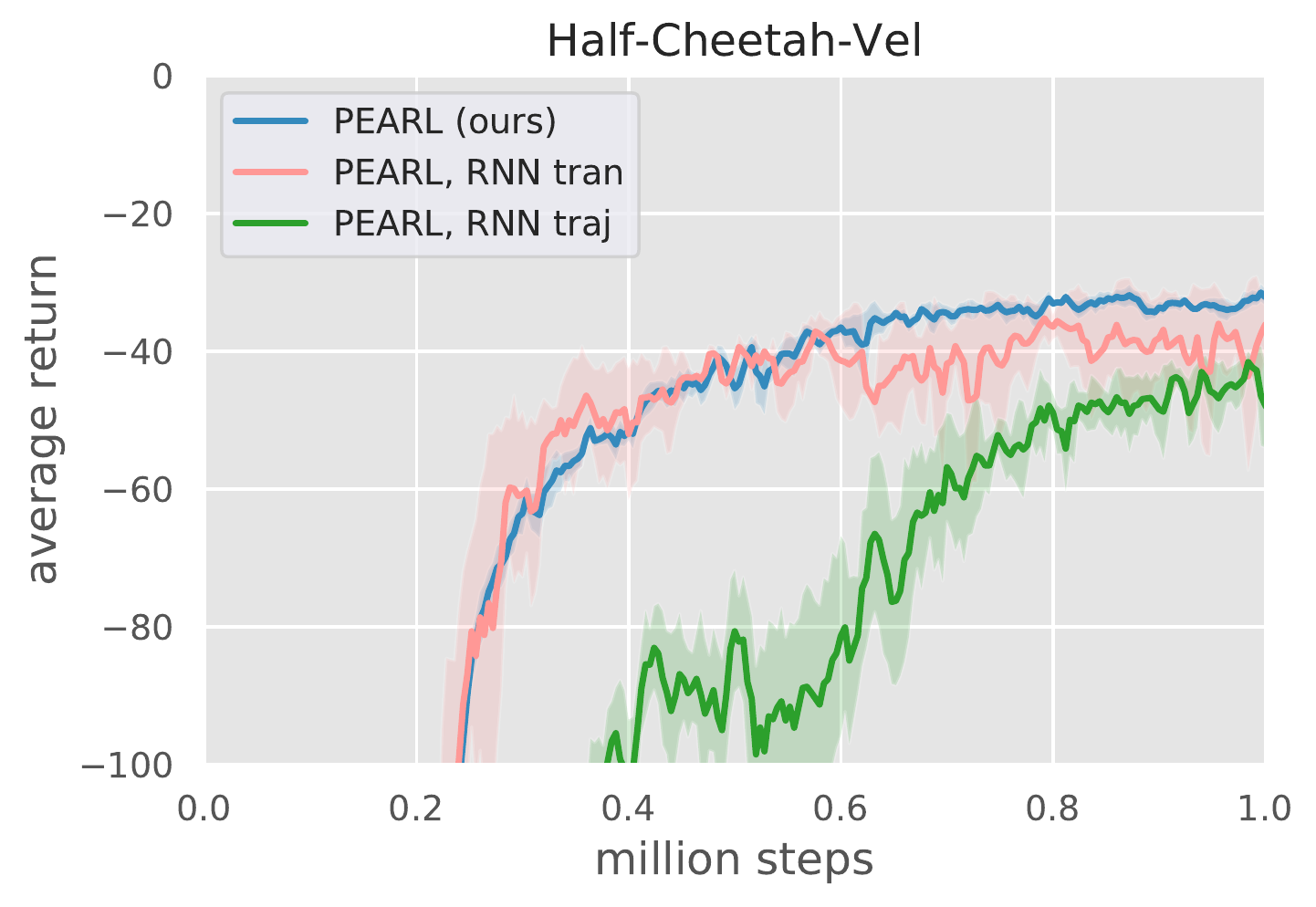}
    \caption{{\bf Recurrent encoder ablation}. We compare our encoder architecture to a recurrent network. We sample context as trajectories rather than unordered transitions. Sampling the RL batch as de-correlated transitions (``RNN tran") fares much better than sampling trajectories (``RNN traj").\vspace{-2mm}}
  \label{fig:rnn-ablation-hc-vel}
\end{figure}

\minisection{Inference network architecture.}
We examine our choice of permutation-invariant encoder for the latent context $\Z$ by comparing it to a conventional choice for encoding MDPs, a recurrent network \cite{duan2016rl, heess2015memory}.
Note that while in Section~\ref{sec:results-perf} we considered a recurrent-based baseline similar to recurrent DDPG \cite{heess2015memory}, here we retain all other features of our method and ablate only the encoder structure.
We backprop through the RNN to 100 timesteps.
We sample the context as full trajectories rather than unordered transitions as in PEARL.
We experiment with two options for sampling the RL batch:
\vspace{-7mm}
\begin{itemize}
    \item unordered transitions as in PEARL (``RNN tran")
    \vspace{-3mm}
    \item sets of trajectories (``RNN traj")
\end{itemize}
\vspace{1mm}
In Figure~\ref{fig:rnn-ablation-hc-vel}, we compare the test task performance in the Half-Cheetah-Vel domain as a function of the number of meta-training samples.
Replacing our encoder with an RNN results in comparable performance to PEARL, at the cost of slower optimization.
However, sampling trajectories for the RL batch results in a steep drop in performance.
This result demonstrates the importance of decorrelating the samples used for the RL objective.

\begin{figure}[t]
  \centering
    \includegraphics[width=0.485\textwidth]{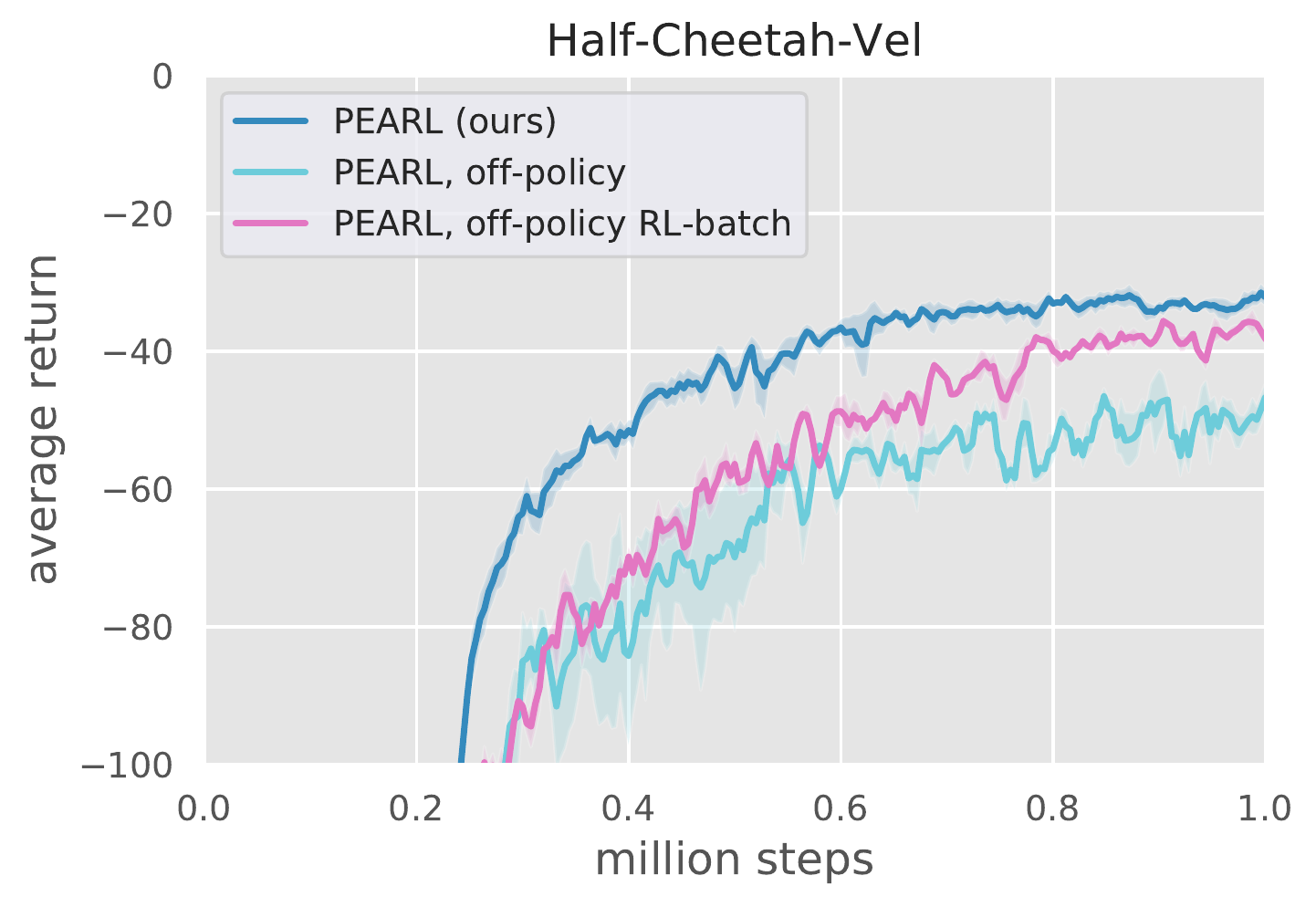}
    \caption{{\bf Context sampling ablation}. PEARL samples context batches of recently collected transitions de-correlated with the batches sampled for RL. We compare to sampling context from the entire history (``off-policy''), as well as using the same sampled batch for the context and the RL batch (``off-policy RL'').}
    \label{fig:data-ablation-hc-vel}
\end{figure}
\minisection{Data sampling strategies.}
In our next experiment, we ablate the context sampling strategy used during training. 
With sampler $\mathcal{S}_\context$, PEARL samples batches of unordered transitions that are \textbf{(1)} restricted to samples recently collected by the policy, and \textbf{(2)} distinct from the set of transitions collected by the RL mini-batch sampler.
We consider two other options for $\mathcal{S}_\context$:
\vspace{-3mm}
\begin{itemize}
    \item sample fully off-policy data from the entire replay buffer, but distinct from the RL batch (``off-policy")
    \item use the same off-policy RL batch as the context (``off-policy RL-batch")
\end{itemize}
\vspace{-3mm}
Results are shown in Figure~\ref{fig:data-ablation-hc-vel}.
Sampling context off-policy significantly hurts performance.
Using the same batch for RL and context in this case helps, perhaps because the correlation makes learning easier.
Overall these results demonstrate the importance of careful data sampling in off-policy meta-RL.

\begin{figure}
  \centering
    \includegraphics[width=0.482\textwidth]{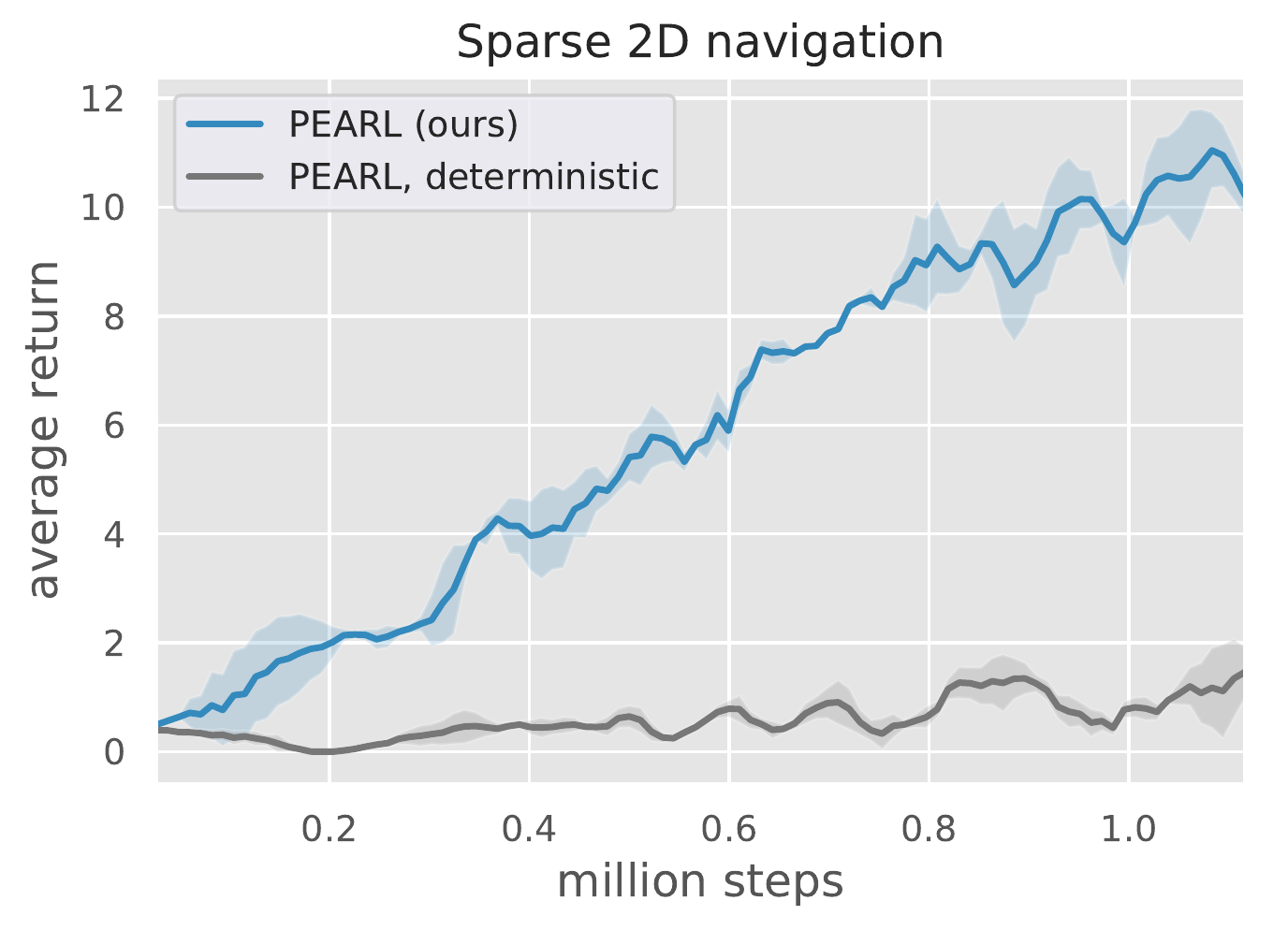}
    \caption{{\bf Deterministic latent context}. We compare PEARL to a variant with deterministic latent context on the sparse reward 2D navigation domain. As expected, without a mechanism for reasoning about uncertainty over tasks, this approach is unable to explore effectively and performs poorly.\vspace{-2.5mm}}
    \label{fig:ablate-prob}
\end{figure}

\minisection{Deterministic context.}
Finally, we examine the importance of modeling the latent context as probabilistic.
As discussed in Section~\ref{sec:context}, we hypothesize that a probabilistic context is particularly important in sparse reward settings because it allows the agent to model a distribution over tasks and conduct exploration via posterior sampling.
To test this empirically, we train a deterministic version of PEARL by reducing the distribution $q_\phi(\z | \context)$ to a point estimate.
We compare probabilistic and deterministic context on the sparse navigation domain in Figure~\ref{fig:ablate-prob}.
With no stochasticity in the latent context variable, the only stochasticity comes from the policy and is thus time-invariant, hindering temporally extended exploration.
As a result this approach is unable to solve a sparse reward navigation task.

\section{Conclusion}
In this paper, we propose a novel meta-RL algorithm, PEARL, which adapts by performing inference over a latent context variable on which the policy is conditioned. 
Our approach is particularly amenable to off-policy RL algorithms as it decouples the problems of inferring the task and solving it, allowing for off-policy meta-training while minimizing mismatch between train and test context distributions.
Modeling the context as probabilistic enables posterior sampling for exploration at test time, resulting in temporally extended exploration behaviors that enhance adaptation efficiency.
Our approach obtains superior results compared to prior meta-RL algorithms while requiring far less experience on a diverse set of continuous control meta-RL domains.

\minisection{Acknowledgements}
We gratefully acknowledge Ignasi Clavera for running the MAML, ProMP, and RL$^2$ baselines.
We thank Ignasi Clavera, Abhishek Gupta, and Rowan McAllister for insightful discussions, and Coline Devin, Kelvin Xu, Vitchyr Pong, and Karol Hausman for feedback on early drafts. This research was supported by NSF IIS-1651843 and IIS-1614653, Berkeley DeepDrive, the Office of Naval Research, ARL DCIST CRA W911NF-17-2-0181, Amazon, Google, and NVIDIA.

\bibliography{ms}
\bibliographystyle{icml2019}


\vspace{5.8in}

\begin{figure*}[t!]
  \centering
    \includegraphics[width=1.0\textwidth]{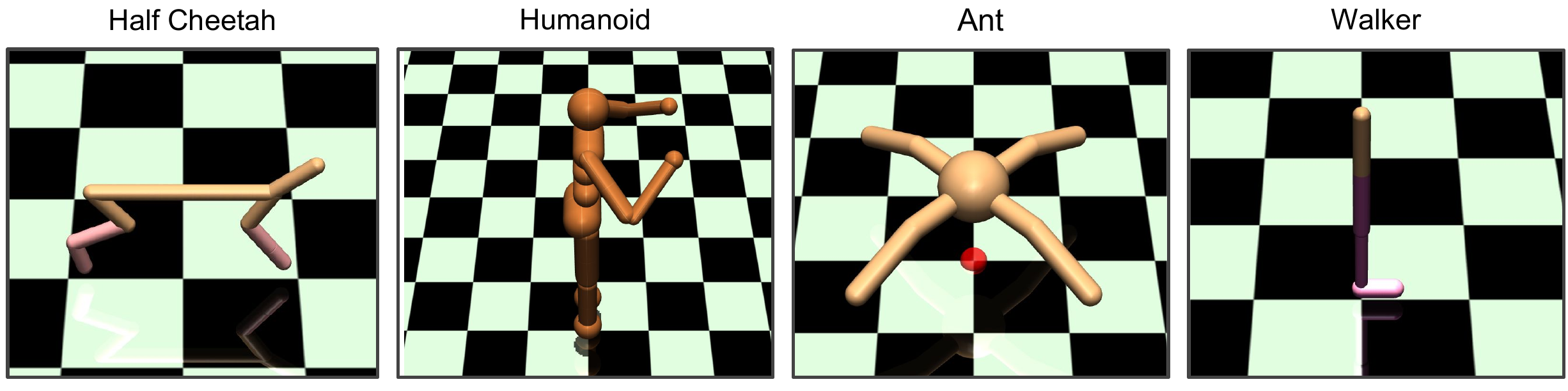}
\vspace{-0.2in}
  \caption{{\bf Continuous control tasks}: left-to-right: the half-cheetah, humanoid, ant, and walker robots used in our evaluation.}
  \label{fig:agents}
\vspace{-0.1in}
\end{figure*}

\begin{figure*}[t!]
  \centering
    \includegraphics[width=1.035\textwidth]{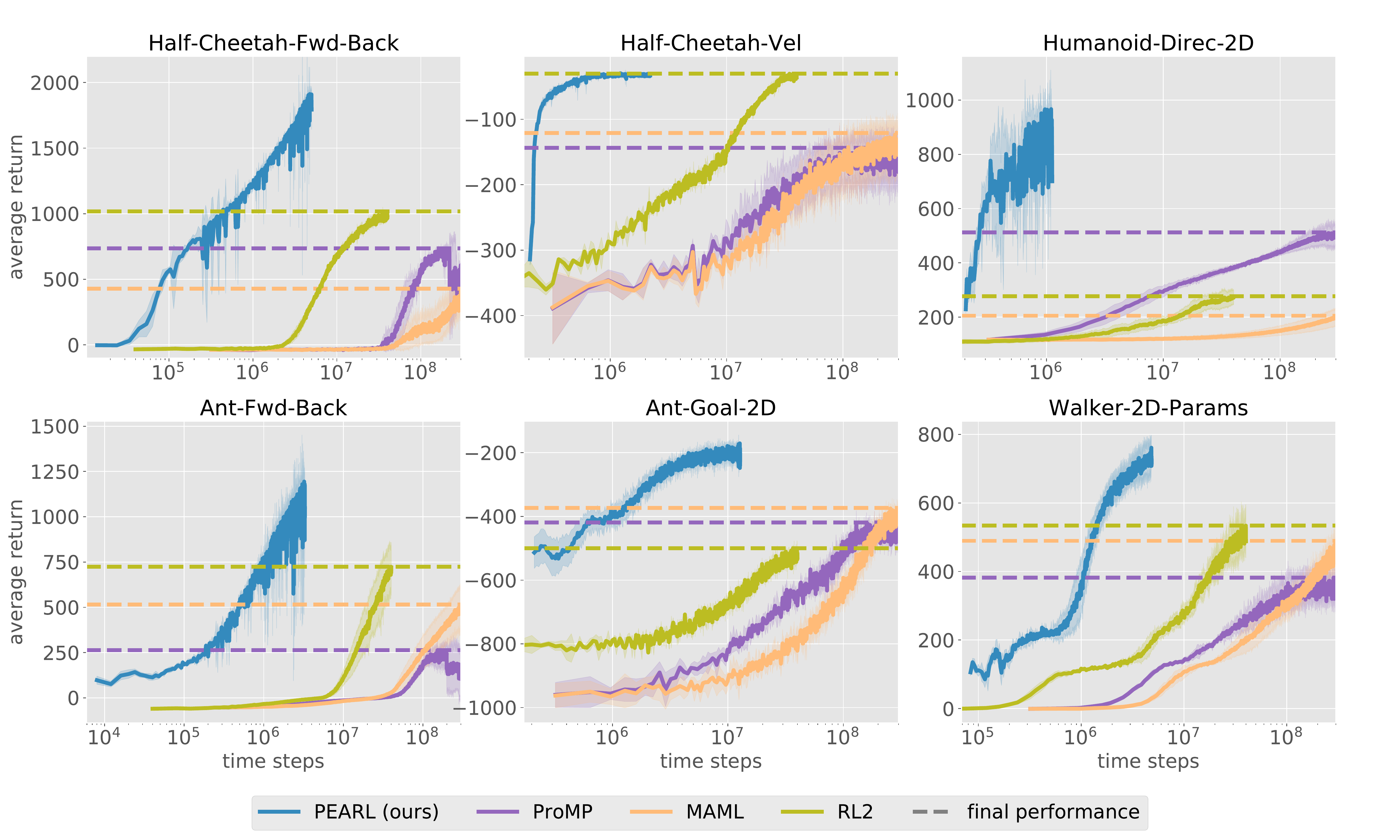}
\vspace{-0.25in}
  \caption{{\bf Meta-learning continuous control}. Test task performance vs. samples collected during \emph{meta-training}. While in the main paper we truncate the x-axis to better illustrate the performance of PEARL, here we plot PEARL against the on-policy methods run for the full number of time steps ($1e8$). PEARL is $20$-$100$ times more sample efficient. Note that the x-axis is in \textbf{log scale}.}
  \label{fig:complete-core-results}
\vspace{-0.1in}
\end{figure*}

\appendix

\section{Experimental Details}
\label{appendix:full-timescale}
The on-policy baseline approaches require many more samples to learn the benchmark tasks. 
Here we plot the same data as in Figure~\ref{fig:core-results} for the full number of time steps used by the baselines, in Figure~\ref{fig:complete-core-results}.
The agents used in these continuous control domains are visualized in Figure~\ref{fig:agents}.
Here we describe each meta-learning domain.
\vspace{-0.1in}
\begin{itemize}
    \item Half-Cheetah-Dir: move forward and backward (2 tasks)
    \item Half-Cheetah-Vel: achieve a target velocity running forward (100 train tasks, 30 test tasks)
    \item Humanoid-Dir-2D: run in a target direction on 2D grid (100 train tasks, 30 test tasks)
    \item Ant-Fwd-Back: move forward and backward (2 tasks)
    \item Ant-Goal-2D: navigate to a target goal location on 2D grid (100 train tasks, 30 test tasks)
    \item Walker-2D-Params: agent is initialized with some system dynamics parameters randomized and must move forward (40 train tasks, 10 test tasks)
\end{itemize}

\end{document}